\documentclass[11pt]{article}

\usepackage{acl}

\usepackage{times}
\usepackage{latexsym}
\usepackage[T1]{fontenc}
\usepackage[utf8]{inputenc}
\usepackage{microtype}
\usepackage{inconsolata}
\usepackage{graphicx}
\usepackage{amsmath}
\usepackage{booktabs}

\newcommand{\lowerbetter}{\ensuremath{(\downarrow)}}
\newcommand{\insection}[2][.]{{\setlength{\parskip}{6pt} \noindent\textbf{#2#1}}}

\title{Statistically Reliable LLM-Based Ranking Evaluation \\ via Prediction-Powered Inference}

\author {
    Abhishek Divekar \\
    Amazon \\
    adivekar@amazon.com
}

\begin{document}
\maketitle
\begin{abstract}
\vspace{-0.25em}
With PRECISE\footnote{Extended abstract; see the full PRECISE paper at: \url{https://doi.org/10.1609/aaai.v40i47.41427}}, we extended Prediction-Powered Inference to produce bias-corrected estimates of ranking evaluation metrics by combining a small human-labeled set with a large LLM-judged set.
PPI is provably unbiased regardless of the LLM judge's error profile.
We make it applicable to hierarchical metrics like Precision@K, where annotations are per-document but the metric is per-query, by reducing the output-space computation from $O(2^{|C|})$ to $O(2^K)$.
On the ESCI benchmark, augmenting 30 human annotations with Claude~3 Sonnet judgments reduces the standard error of Precision@4 estimates from 4.45 to 3.50 (a 21\% relative reduction).
In a production system, our framework correctly identified the best of three system variants from 100 human labels and 2 hours of domain-expert annotation; A/B testing confirmed this ranking with +407~bps in daily sales.
\vspace{0.0em}
\end{abstract}

\section{Introduction}
\vspace{-0.25em}

Human evaluation is expensive, yet smaller labeled sets produce wide confidence intervals that cannot distinguish genuine system improvements from noise. LLM-as-a-Judge approaches attempt to address this, but carry systematic biases that distort evaluation metrics when used as substitutes for human annotation \citep{chen-etal-2024-humans}.

Most prior work addresses this tension by building better judges through prompt engineering, fine-tuning, or multi-agent debate.
We take an orthogonal approach: accept that LLM judges are biased and \emph{correct for the bias statistically}.
Our framework extends Prediction-Powered Inference \citep[PPI;][]{angelopoulos2023ppi}, a semi-supervised estimation method that combines a small gold set (human labels) with a large LLM-annotated set.
The gold set measures the judge's systematic error and corrects for it.
The resulting estimate is provably unbiased, and each additional LLM-judged example reduces the variance of the metric estimate without introducing new bias.

A challenge arises for metrics that aggregate granular judgments into a higher-level score: 
for Precision@K, human annotations are collected per-document but the metric is computed per-query.
Standard PPI cannot handle this granularity mismatch.
We resolve this through a sparse reformulation of the output space (\S\ref{sec:method}), and validate on a public benchmark and a production A/B test (\S\ref{sec:results}).

\section{Method}
\label{sec:method}
\vspace{-0.25em}

Let $\mathcal{D}_g = \{(x^{(i)}_g, y^{(i)}_g)\}_{i=1}^{n}$ be a small gold set with human labels and $\mathcal{D}_u = \{x^{(i)}_u\}_{i=1}^{N}$ a large set ($N \gg n$) annotated by an LLM $M$. The PPI++ estimator \citep{angelopoulos2024ppiefficient} evaluates:
\begin{align}
\hat{\mu}_{\text{\scriptsize PPI}} &= \underbrace{\frac{\lambda}{N}\!\sum_{i=1}^{N} \tilde{\mu}_u^{(i)}}_{\text{LLM-based estimate}} + \underbrace{\frac{1}{n}\!\sum_{i=1}^{n}\!\Big[\phi_i - \lambda\, \tilde{\mu}_g^{(i)}\Big]}_{\text{bias correction}}
\label{eq:ppi}
\end{align}
where $\phi_i$ is the human-grounded metric on the $i$-th gold query, and $\tilde{\mu}_u^{(i)}, \tilde{\mu}_g^{(i)}$ are the LLM-based metric estimates obtained by marginalizing over the judge's output distribution.
The parameter $\lambda \in [0,1]$ is tuned to minimize the variance of $\hat{\mu}_{\text{\scriptsize PPI}}$; the estimator remains unbiased for any $\lambda > 0$.

The bias-correction term (second summand) measures how the LLM judge deviates from human ground truth on the gold set, then subtracts this deviation from the LLM-only estimate.
When the LLM is well-calibrated, setting $\lambda \simeq 1$ allows the full unlabeled set to drive variance reduction.
When the LLM is heavily biased, we can shrink $\lambda \simeq 0$ and the estimator relies on gold estimates.

\vspace{-0.5em}
\paragraph{Hierarchical metrics.}
For Precision@K, annotations are at the query-document level but metrics are calculated per-query.
The naive PPI output space is $\{0,1\}^{|C|}$ (one binary relevance variable per corpus document), which is computationally intractable when $|C|$ is in the millions.

We observe that Precision@K depends only on the top-$K$ retrieved documents; thus, it reduces to a scaled dot product over sparse vectors: $\phi(\hat{y}, y) = \hat{y}^\top y / K$.
Because only $K$ positions contribute, the probability mass of all non-retrieved documents collapses into a single weight on the all-zero $K$-vector.
This reduces the output space to $\{0,1\}^K$.

For each query, the LLM judge provides per-document relevance probabilities $\tilde{p}'(d_k)$ for the $k$-th ranked result.
We form a joint distribution over $K$-length binary vectors assuming conditional independence across documents:
\vspace{-0.5em}
\begin{align}
\tilde{p}(y) = \prod_{k=1}^{K} \tilde{p}'(d_k)^{y_k}(1{-}\tilde{p}'(d_k))^{(1{-}y_k)}
\label{eq:joint}
\end{align}

\vspace{-0.5em}

where $y_k$ is the $k$-th element of $y \in \{0,1\}^K$.
The LLM-based estimates $\tilde{\mu}^{(i)}$ in Eq.~\ref{eq:ppi} are then computed by summing $\phi(\hat{y}, y) \cdot \tilde{p}(y)$ over all $2^K$ vectors.
For typical $K \leq 10$ this sum is tractable.

\section{Results}
\label{sec:results}
\vspace{-0.25em}

We validate on the ESCI retrieval benchmark \citep{reddy2022shopping} using Claude~3 Sonnet and Haiku as LLM judges, with $n{=}30$ gold annotations and $N{=}60{,}000$ unlabeled queries.

\begin{figure}[t]
    \centering
    \includegraphics[width=\columnwidth]{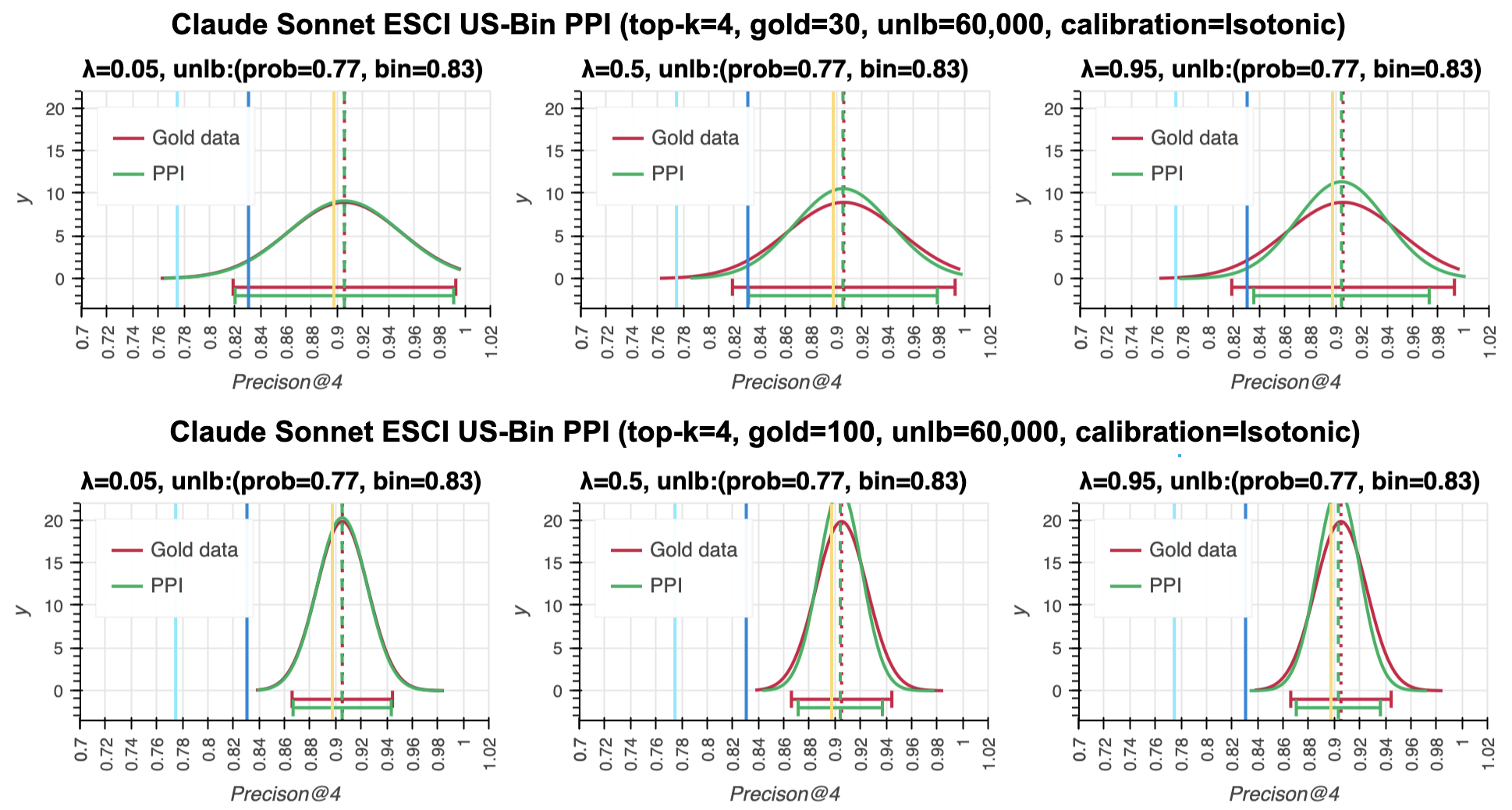}
    \caption{Sampling distributions for Precision@4 on ESCI ($N{=}60$K, Claude~3 Sonnet). Top: $n{=}30$; bottom: $n{=}100$. PPI (green) is tighter than gold-only (red) and centered on the true value (yellow dashed). LLM-only estimates (cyan, blue) are biased.}
    \label{fig:ppi_results}
    \vspace{-0.75em}
\end{figure}

\begin{table}[t]
  \centering
  \small
  \begin{tabular}{lrrr}
    \toprule
    \textbf{Estimator} & \textbf{Bias} \lowerbetter & \textbf{Std.\ Err.} \lowerbetter & \textbf{Cost} \\
    \midrule
    Gold only ($n{=}30$) & 1.04 & 4.45 & \textemdash \\
    \midrule
    + Claude 3 Sonnet & 0.70 & \textbf{3.50} & \$946 \\
    + Claude 3 Haiku & \textbf{0.29} & 3.86 & \$79 \\
    \bottomrule
  \end{tabular}
  \vspace{-0.75em}
  \caption{Precision@4 estimation on ESCI ($n{=}30$ gold, $N{=}60$K LLM-judged). Sonnet reduces standard error from 4.45 to 3.50 (21\% relative reduction); Haiku achieves the lowest bias at 12$\times$ lower cost.}
  \label{tab:cost}
  \vspace{-1.75em}
\end{table}

\vspace{-0.5em}
\paragraph{Variance reduction and cost.}
Table~\ref{tab:cost} shows that PPI with Sonnet reduces standard error from 4.45 to 3.50 (\textbf{-21\% relative}) while maintaining low bias (0.70 vs.\ 1.04 for gold-only).
Haiku achieves comparable quality (SE:~3.86, bias:~0.29) at 12$\times$ lower inference cost.
Figure~\ref{fig:ppi_results} illustrates why: the PPI sampling distribution (green) is narrower than gold-only (red) because the LLM signal reduces variance, and it stays centered on the true value (yellow dashed) because the bias-correction term removes the judge's systematic error.
Separately, we found the framework plateaus at a 100$\times$ unlabeled-to-gold ratio: $N=3,000$ LLM queries provide nearly identical standard error to $N=60,000$ with $n=30$ labelled examples.

\vspace{-0.5em}
\paragraph{Production A/B test.}
In a production search system, our Precision@K formulation ranked three system variants (C, T1, T2) using $n{=}100$ human labels and $N{=}8{,}400$ LLM judgments, produced in 2~hours of expert annotation.
The predicted ranking (T1~$>$~T2~$>$~Control) was confirmed by A/B testing: T1 yielded +407~bps in daily sales and +571~bps in click-through rate.
Without PPI correction, LLM-only estimates could not distinguish between variants, because systematic upward bias inflated all estimates similarly; introducing semi-supervised estimation restored discriminative power by correcting for this bias.

Though we validate it on Precision@K, the hierarchical formulation applies in principle to any metric that aggregates fine-grained judgments (e.g., per-claim factuality, per-turn dialogue quality).

\section{Future Work}
\vspace{-0.25em}
Several promising directions remain for future work. We describe a few of them.

\insection{Synthetic covariates} An over-reliance on human labels is a major drawback of our estimation method. LLM-generated synthetic datasets can begin from fixed but gold labels and provide good textual covariates, which may nevertheless be usable for estimation \cite{yu2023large, kowshik-etal-2024-corrsynth}.

\insection[]{Doubly robust estimation} \cite{oosterhuis2023doubly} shares a theoretical grounding with LLM bias and could offer a pathway toward real-time, bias-corrected metric inference. Adopting this paradigm would enable online evaluation for our method.

\insection{Multiple Judges} A complementary line of work aggregates verdicts across several LLM judges, which may match human ratings more closely than a single model \cite{zheng2024judging}. However, the alternative of folding several rubrics into one judge prompt and tuning it jointly turns out to be brittle in practice \cite{darshan2026gradientscollidefailuremodes}. A natural extension of our method, then, is to adopt multi-objective optimization procedures, rather than depending on a single all-purpose evaluator.

\insection{Agentic critics} Agentic systems increasingly rely on LLM-based critics to score and refine their own outputs \cite{yuksekgonul2025optimizing, rudman2026vesta}, yet these critics inherit the same biases PRECISE corrects. Extending our approach to produce calibrated critic signals from minimal human labels is a promising direction for more reliable agent optimization.

\section*{Acknowledgments}
\vspace{-0.5em}
Anirban Majumder contributed to the original version of this work \citep{Divekar_Majumder_2026}.

\section*{Ethics Statement}
\vspace{-0.5em}

All user queries in the production A/B test were anonymized to remove personally identifiable information before being processed by LLM judges or human annotators.
Human annotation was performed by domain experts during normal working hours as part of their regular responsibilities; no crowdworker labor was used.
The framework reduces (but does not eliminate) the need for human annotation; it is not intended to replace human evaluation entirely, but to make small human annotation budgets go further.

\section*{Limitations}
\vspace{-0.5em}

Our framework has three limitations.
First, we have validated the hierarchical PPI extension only on Precision@K for retrieval; generalization to other hierarchical metrics (e.g., per-claim factuality, per-turn dialogue quality) remains untested.
Second, the conditional independence assumption across documents in Eq.~\ref{eq:joint} may not hold when relevance of one document depends on the presence of another (e.g., diversity-sensitive ranking); relaxing this assumption is left to future work.
Third, the framework requires a small gold set from the same distribution as the unlabeled set; distribution shift between the gold and unlabeled queries (e.g., from temporal drift) could degrade the bias correction.

\vspace{-1.5em}
\bibliography{custom-GEM}

\end{document}